\documentclass[11pt,letterpaper]{article}

\usepackage[T1]{fontenc}
\usepackage{lmodern}

\usepackage[margin=1in]{geometry}
\usepackage{microtype}

\usepackage{amsmath,amssymb}
\usepackage{graphicx}

\usepackage{enumitem}

\usepackage[hidelinks,breaklinks=true]{hyperref}
\usepackage{cleveref}

\title{Time, Causality, and Observability Failures\\
in Distributed AI Inference Systems}

\author{Ankur Sharma\textsuperscript{1} \and
Deep Shah\textsuperscript{2} \and
David Lariviere\textsuperscript{3} \and
Hesham ElBakoury\textsuperscript{4} \\[6pt]
\textsuperscript{1}\texttt{ankur.sharma@ocproject.net} \quad
\textsuperscript{2}\texttt{deepshah146@gmail.com} \\
\textsuperscript{3}\texttt{davidl@illinois.edu} \quad
\textsuperscript{4}\texttt{helbakoury@gmail.com}}
\date{April 2026}

\begin{document}

\maketitle

\begin{abstract}
Distributed AI inference pipelines rely heavily on timestamp-based
observability to understand system behavior. This work demonstrates that
even small clock skew between nodes can cause observability to become
causally incorrect while the system itself remains functionally correct
and performant. We present controlled experiments on a multi-node AI
inference pipeline, where clock skew is introduced at a single stage.
Results show that no violations are observed under synchronized
conditions and up to 3\,ms skew, while clear causality violations emerge
by 5\,ms. Despite this, system throughput and output correctness remain
largely unaffected. We further observe that violation behavior is not
strictly static. In longer runs, negative span rates may stabilize or
decrease over time, indicating that effective skew evolves due to
relative clock drift between nodes. Experiments were conducted using
Kafka and ZeroMQ transports, with consistent results across both. Aeron
is under active exploration but is not yet included in the completed
validation set. These findings suggest that observability correctness
depends not only on system functionality but also on precise time
alignment, and that timing must be treated as a first-class concern in
distributed AI systems. These results highlight that observability
correctness depends not only on system functionality but also on precise
time alignment.
\end{abstract}

\section{Introduction and Motivation}
\label{sec:intro}

AI inference pipelines have rapidly evolved from monolithic services
into distributed systems composed of multiple stages, often spanning
hosts, racks, regions, and administrative domains. Typical pipelines
include request producers, preprocessing stages, inference services,
postprocessing stages, and downstream consumers connected by
message-oriented middleware. To operate, debug, audit, and scale these
systems, operators rely heavily on timestamp-based observability: logs,
traces, metrics, and event timelines ordered by wall-clock time.

Implicit in this approach is a foundational assumption: clocks across
the system are sufficiently synchronized such that timestamps remain
safe to reason with. In practice, this assumption is rarely questioned.
Small clock skew is treated as benign, tolerated if it does not cause
visible performance degradation or functional errors.

This paper challenges that assumption. We show that distributed AI
inference systems can continue operating correctly producing valid
outputs at stable throughput while silently losing the ability to reason
about cause and effect. In such a state, the system's internal timeline
becomes logically inconsistent, audit trails become unreliable, and
post-incident reconstruction becomes impossible, all without triggering
conventional alarms.

The central question we address is not whether clock skew exists --- it
inevitably does --- but what happens when systems continue to reason
about causality as if timestamps remain perfect once skew exists. Our
results demonstrate that the answer is silent observability failure.

This work was conducted as part of the Unified Intelligent Infrastructure
workstream at the Open Compute Project
(OCP).\footnote{\url{https://www.opencompute.org/w/index.php?title=Unified_Intelligent_Infrastructure}}

\subsection*{Contributions}
This work makes the following contributions:
\begin{itemize}[leftmargin=*]
  \item An experimental demonstration that small clock skew is
    sufficient to silently break timestamp-based causal observability in
    distributed AI inference pipelines while functional behavior remains
    correct.
  \item Identification of an onset region between 3\,ms and 5\,ms skew,
    where causal inconsistencies begin to appear while system throughput
    remains stable.
  \item Observation that causality violations can evolve dynamically
    over time, including partial self-recovery behavior likely driven by
    relative clock drift between nodes.
  \item Introduction of an explicit, binary causality health signal that
    allows systems to detect and surface loss of causal trust at
    runtime.
  \item An analysis of real-world implications across multi-tenant AI
    platforms, autonomous agents, regulated decision systems, and
    incident response workflows.
  \item Validation that the observed failure behavior is consistent
    across Kafka and ZeroMQ transports, indicating that the issue is not
    tied to a specific messaging system.
\end{itemize}

\section{Foundations: Causality, Observability, and Time in Distributed AI Systems}
\label{sec:foundations}

\subsection{Causality in Distributed Systems}
Causality describes the relationship between events where one event
influences or precedes another. In distributed systems, causality is
not directly observable and must be inferred. Lamport~\cite{lamport1978}
formalized this concept by defining the happens-before relationship,
which provides a partial ordering of events without relying on physical
time. In practice, however, many systems approximate causality using
wall-clock timestamps for convenience and scalability.

\subsection{Observability and Inference of System Behavior}
Observability refers to the ability to infer the internal state and
behavior of a system from external signals such as logs, metrics, and
traces~\cite{birman,dapper,otel,burgess2019}. Modern observability stacks
implicitly rely on timestamp ordering to reconstruct execution
timelines, perform root-cause analysis, and verify correctness. When
timestamps are inconsistent, observability does not fail loudly;
instead, it produces contradictory or misleading narratives about
system behavior.

\subsection{Time Synchronization as an Assumption}
Time synchronization protocols aim to bound clock error across
nodes~\cite{ntp,ptp1588}, but perfect synchronization is unattainable.
Oscillator drift, synchronization jitter, transient loss of time
sources, and system-level effects introduce unavoidable skew.
Distributed AI systems often assume that as long as clocks are within
nominal error bounds, timestamp-based reasoning remains valid. This
paper examines the consequences of that assumption.

\section{System Model and Assumptions}
\label{sec:model}

We model a distributed AI inference pipeline as a linear sequence of
stages:
\begin{enumerate}[leftmargin=*]
  \item Request production
  \item Preprocessing
  \item Inference
  \item Postprocessing
  \item Observation and monitoring
\end{enumerate}

Each stage executes on a separate host with its own local clock.
Stages communicate via a reliable messaging substrate that preserves
message ordering and delivery but does not impose global time
semantics.

We assume:
\begin{itemize}[leftmargin=*]
  \item No message loss.
  \item Correct functional behavior of all components.
  \item Stable throughput.
  \item Bounded but non-zero clock skew.
\end{itemize}

The focus of this work is strictly on observability correctness, not
functional correctness or performance optimization.

In later experimental runs, we explicitly re-validated inter-node clock
alignment before each run. This became necessary after observing that
uncontrolled infrastructure time offsets could become large enough to
dominate the intended application-level skew injection.

We do not consider adversarial clock manipulation; skew arises from
benign and unavoidable system effects.

\section{Experimental Methodology}
\label{sec:method}

\subsection{Pipeline Implementation}
We implement a representative AI inference pipeline using commodity
infrastructure and open-source components. A real large language model
is used at the inference stage to ensure realistic compute behavior and
token generation patterns. The messaging substrate and monitoring stack
are representative of widely deployed production systems. While the
core analysis is framed generically, we include a concrete reference
implementation to provide experimental context and reproducibility.

\begin{figure}[!htbp]
  \centering
  \includegraphics[width=0.45\linewidth]{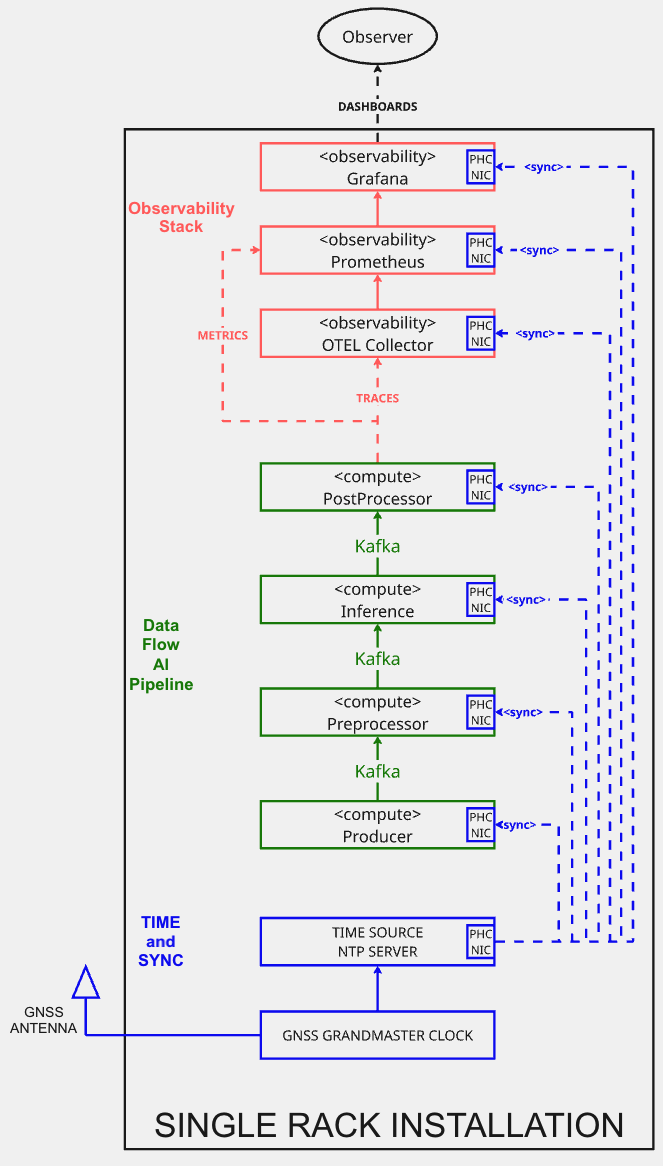}
  \caption{Distributed AI inference pipeline used in our experiments.
    Each stage executes on a separate host with an independent local
    clock and communicates via a message-oriented substrate.
    Observability is derived by comparing timestamps emitted at
    different stages rather than by direct causal signaling. This
    structure reflects common production AI inference deployments and
    makes explicit the reliance on timestamp-based reasoning across
    independently clocked components.}
  \label{fig:pipeline}
\end{figure}

\subsubsection{Reference Implementation}
To ground the experiments and enable reproducibility, we evaluate a
concrete reference implementation of the distributed AI inference
pipeline. The initial implementation used Kafka to establish a stable
distributed baseline and support the first phase of experiments. A
second implementation used ZeroMQ as a brokerless transport to reduce
middleware influence and simplify direct stage-to-stage communication.
The same qualitative observability failure mode was observed across
both Kafka and ZeroMQ. Aeron is under active exploration as a
lower-latency transport alternative, but due to current Python binding
and integration challenges, it is not yet included in the completed
validation set reported in this paper.

\begin{description}[leftmargin=*,style=unboxed]
  \item[Hardware.] The experiments are conducted on a small cluster of
    commodity x86 servers located within a single rack and connected
    via a top-of-rack switch. Each pipeline stage executes on a
    separate physical host. The servers use multi-core CPUs with
    sufficient memory to support inference workloads without swapping
    or resource contention. No specialized accelerators are required
    for the experiments presented.
  \item[Operating System and Kernel.] All hosts run a modern Linux
    distribution with a recent general-purpose kernel. Default kernel
    scheduling and interrupt handling are used, without CPU isolation,
    IRQ pinning, or real-time kernel modifications. This choice
    reflects typical production deployments and avoids introducing
    specialized tuning that could obscure baseline behavior.
  \item[Model Class.] The inference stage uses a CPU-based large
    language model representative of contemporary transformer-style
    generative models. Deterministic decoding is enabled to ensure
    repeatability across runs. The model operates in a request-driven
    inference mode, generating a bounded number of output tokens per
    request. The specific model architecture and size are not
    fundamental to the observed causality behavior; rather, the model
    serves to introduce realistic compute latency and token generation
    patterns.
  \item[Messaging Substrate Class.] The completed experiments in this
    paper use two transport configurations: Kafka and ZeroMQ. Kafka was
    used in the initial implementation to establish a stable baseline,
    while ZeroMQ was introduced to evaluate the same pipeline behavior
    in a simpler brokerless configuration. Aeron remains under active
    exploration and is therefore not included in the completed
    experimental validation set presented here.
  \item[Monitoring and Observability Stack.] Metrics and logs are
    collected using a standard time-series monitoring stack commonly
    deployed in cloud-native systems. Observability relies on
    timestamped metrics and event logs collected independently at each
    stage. No external clock correction or post-hoc timestamp
    reconciliation is performed, reflecting typical operational
    practice.
  \item[Clock Synchronization and Skew Injection.] Before each
    experiment, hosts are manually synchronized to a common time source
    to establish a tight baseline. Continuous synchronization daemons
    are disabled during controlled runs to prevent background correction
    from interfering with the experiment. Skew is then introduced at
    the application level on the inference stage only, allowing direct
    isolation of the effect on observability while preserving system
    functionality.
  \item[Representativeness.] The reference implementation is intended
    to be representative rather than exhaustive. While specific
    hardware, software, and tooling choices may vary across
    deployments, the essential properties studied --- independent
    clocks, timestamp-based observability, and forward-only message
    delivery --- are common across modern distributed AI inference
    systems. The causality and observability behaviors observed are
    therefore expected to generalize beyond this specific
    implementation.
\end{description}

\subsection{Clock Baseline}
Before each experiment, all participating hosts are manually
synchronized to a common time source. In the final controlled runs,
measured inter-node offsets were reduced to a tight baseline suitable
for experimentation. Continuous synchronization daemons were disabled
during runs to prevent background clock corrections from altering the
intended skew conditions.

During early experimentation, we observed that automatic time
synchronization services could reintroduce large host offsets even after
manual correction. In some cases, measured offsets exceeded 600\,ms
relative to the reference clock, far beyond the observed failure onset
region.

To ensure experimental validity, automatic synchronization services were
disabled during runs, and clocks were manually aligned prior to each
experiment. After correction, inter-node offsets were maintained within
a sub-millisecond range, providing a stable baseline for controlled skew
injection.

This baseline procedure became especially important after early
experimentation revealed that uncontrolled infrastructure time offsets
could become very large, in some cases reaching hundreds of
milliseconds. Those offsets were large enough to exceed the observed
observability failure onset by orders of magnitude, making explicit
baseline verification a necessary part of the methodology.

\subsection{Skew Injection}
Clock skew is injected at the inference stage by offsetting
application-level timestamps. The operating system clock is not
modified. In the updated experimental phase, skew values were tested at
0\,ms, 1\,ms, 2\,ms, 3\,ms, 5\,ms, 10\,ms, and 50\,ms, with particular
focus on the 1--5\,ms region to better characterize the onset of
causality violations. Injecting skew at the inference stage isolates
the effect cleanly while preserving functional correctness and stable
throughput.

\subsection{Metrics and Violation Detection}
The postprocessing stage computes observability metrics, including:
\begin{itemize}[leftmargin=*]
  \item Total tokens processed.
  \item Negative timing spans.
  \item Violation rate.
\end{itemize}

A negative span is recorded when timestamps imply impossible causal
relationships, such as events being received before they were sent. We
refer to such contradictions as negative timing spans, indicating loss
of timestamp-based causal correctness. The total negative-span count
is an aggregate count of detected causal contradictions and should not
be interpreted as a one-to-one count of failed requests or failed
tokens.

In practice, the dominant negative timing span detected in our
experiments compares the inference stage's token send timestamp with
the postprocessing stage's receive timestamp. The inference timestamp
is subject to injected skew, while the postprocessor receive timestamp
reflects the local clock at the observation stage, which is not
skew-adjusted. As a result, the relevant $\Delta t_{\min}$ corresponds
primarily to message transit and buffering latency rather than
inference execution time. This distinction explains why millisecond-scale
skew can immediately yield violations despite inference latencies on
the order of hundreds of milliseconds.

While the postprocessor computes multiple consistency checks, the
dominant violation observed in our experiments corresponds to the edge
between inference token send timestamps and postprocessor receive
timestamps. Future work will decompose violation counts further by
causal edge to distinguish which relationships are most sensitive to
skew under different operating conditions.

We additionally introduce a binary metric, \texttt{causality\_health},
defined as:
\begin{itemize}[leftmargin=*]
  \item $1$ if no violations occur within a rolling 30-second window.
  \item $0$ otherwise.
\end{itemize}
This metric allows the system to explicitly declare loss of causal
trust. The 30-second window is chosen to balance responsiveness with
stability, aligning with typical observability and alerting intervals
rather than representing a fundamental system constant.

\section{Experimental Results}
\label{sec:results}

This section presents experimental results demonstrating the impact of
injected clock skew on causal observability. Each experiment compares a
zero-skew baseline against controlled skew values while holding
workload and functional pipeline behavior stable. The updated results
also capture newer observations from Kafka and ZeroMQ runs, including
onset behavior in the 3--5\,ms region and time-varying violation
behavior in longer runs.

\subsection{\texorpdfstring{Baseline (0\,ms Skew)}{Baseline (0 ms Skew)}}
With no injected skew in the synchronized baseline runs:
\begin{itemize}[leftmargin=*]
  \item Throughput remains stable.
  \item No negative spans are observed.
  \item \texttt{causality\_health} remains $1$.
\end{itemize}
These baseline runs establish the reference condition for the rest of
the paper: functional correctness, stable throughput, and causally
valid observability all hold simultaneously when clocks are properly
aligned.

\begin{figure}[!htbp]
  \centering
  \includegraphics[width=0.75\linewidth]{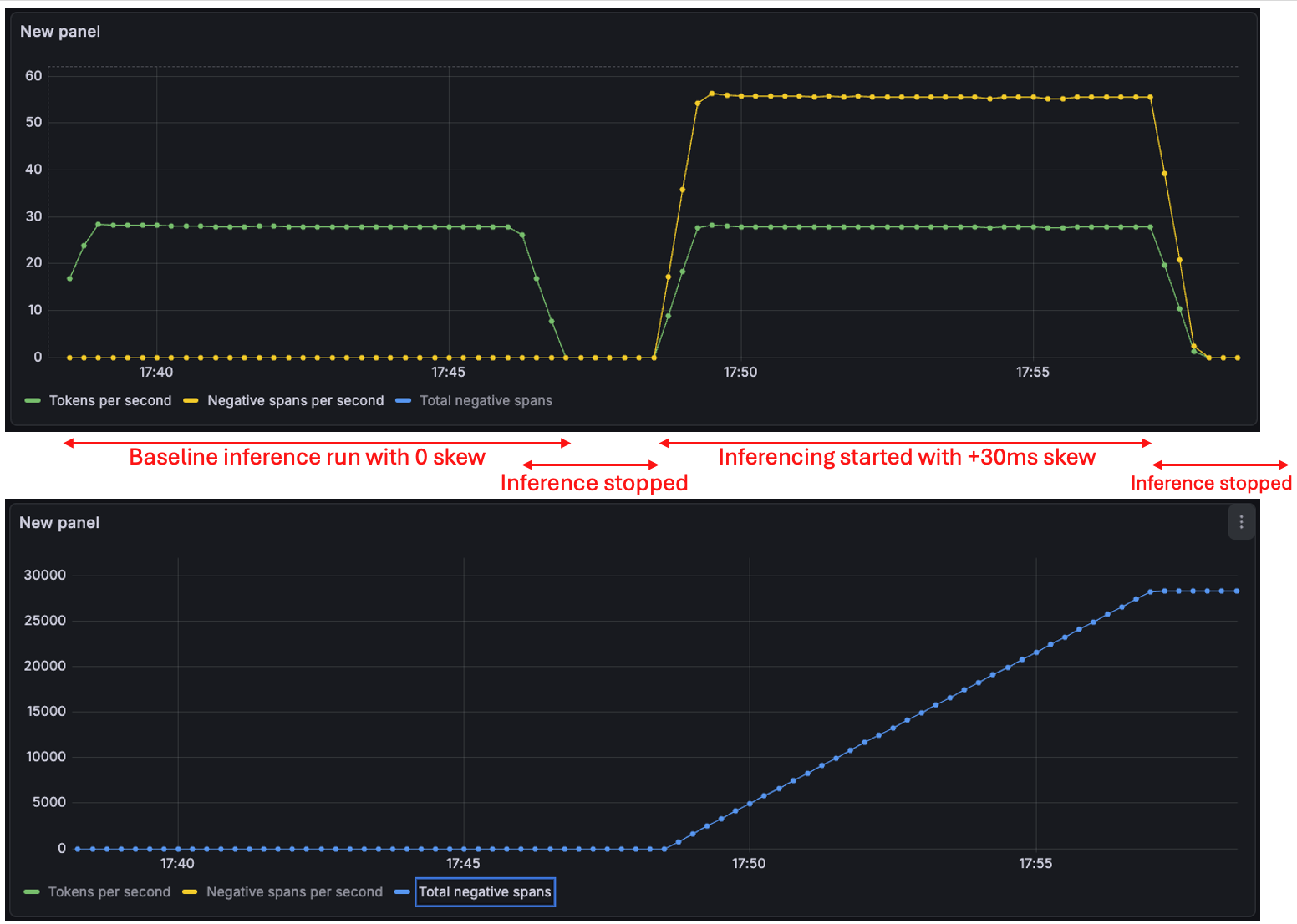}
  \caption{Throughput and violations under zero and non-zero skew.
    Throughput remains stable across both runs, demonstrating that
    functional execution and performance are largely unaffected.
    However, in the skewed run, negative timing spans emerge even
    though throughput stays healthy.}
  \label{fig:throughput}
\end{figure}

\begin{figure}[!htbp]
  \centering
  \includegraphics[width=0.85\linewidth]{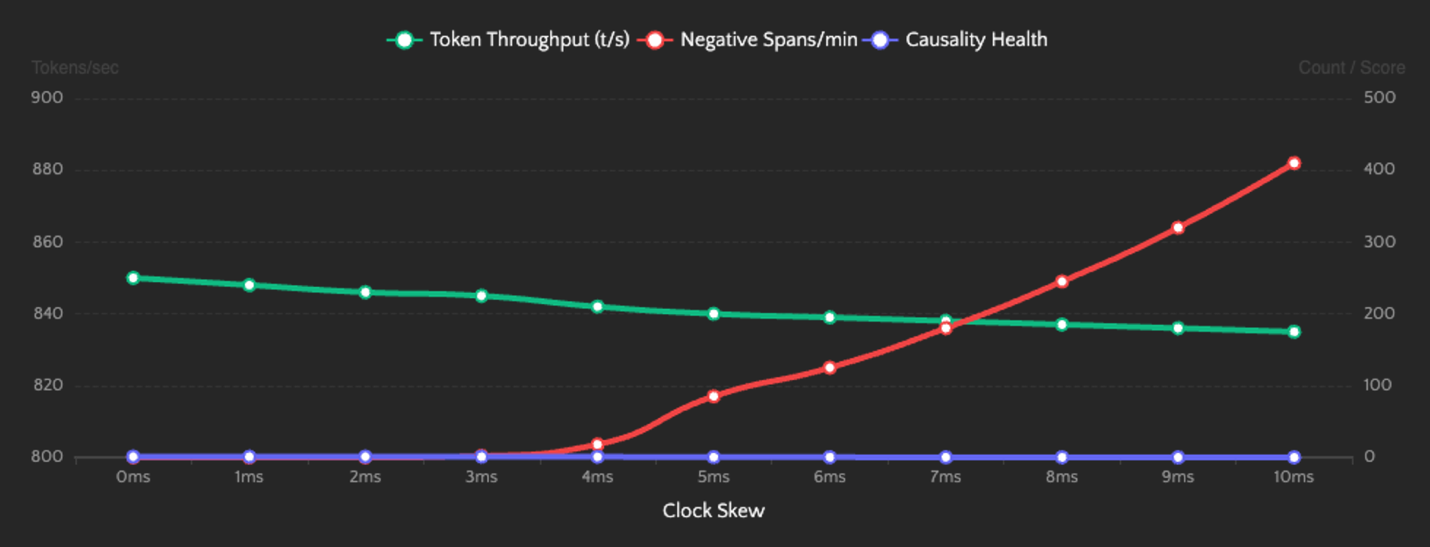}
  \caption{System performance versus observability health. This
    illustrates the central result of the paper: observability can
    become causally incorrect before the system itself shows
    functional failure.}
  \label{fig:perf-vs-health}
\end{figure}

Figures~\ref{fig:throughput} and~\ref{fig:perf-vs-health} compare
system behavior under zero skew and injected clock skew. Throughput
remains stable across both runs, demonstrating that functional
execution and performance are largely unaffected. However, in the
skewed run, negative timing spans emerge even though throughput stays
healthy. This illustrates the central result of the paper: observability
can become causally incorrect before the system itself shows functional
failure.

\begin{figure}[!htbp]
  \centering
  \includegraphics[width=0.55\linewidth]{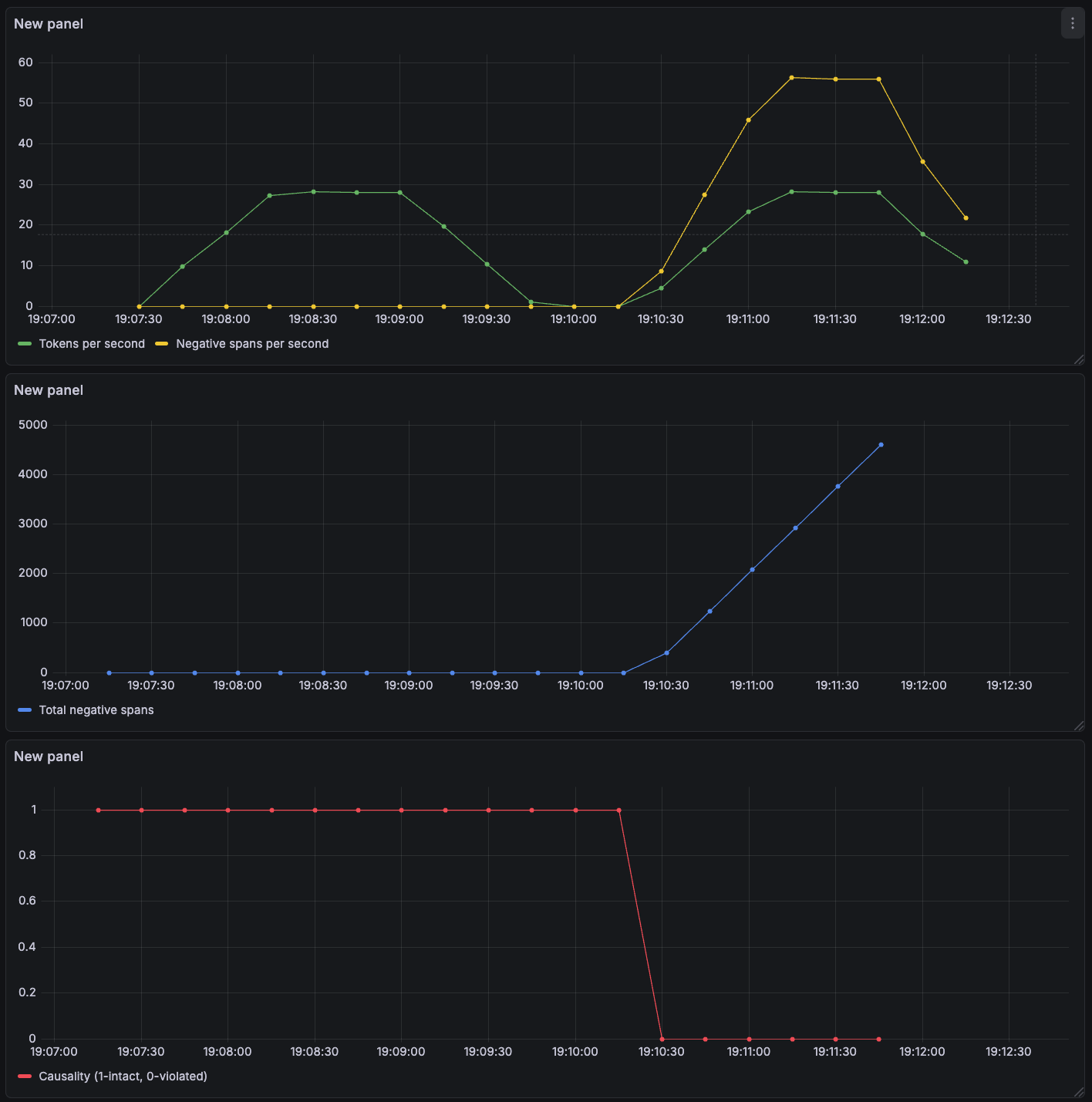}
  \caption{Causality health under skew.}
  \label{fig:causality-health}
\end{figure}

\begin{figure}[!htbp]
  \centering
  \includegraphics[width=0.85\linewidth]{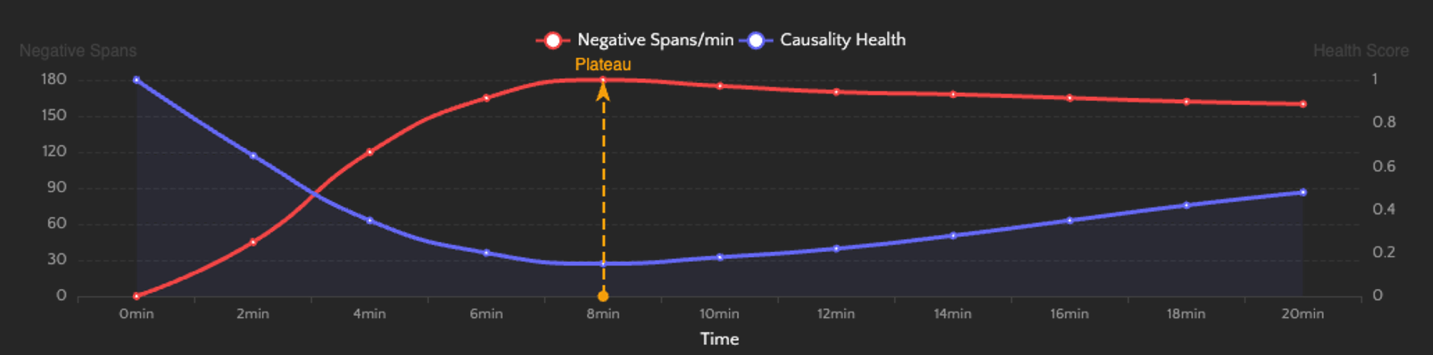}
  \caption{Self-recovery pattern over time.}
  \label{fig:recovery}
\end{figure}

Figures~\ref{fig:causality-health} and~\ref{fig:recovery} show the
\texttt{causality\_health} signal over time. Under zero skew, the
signal remains at $1$, indicating intact causal observability. When
skew is introduced above the onset region, \texttt{causality\_health}
transitions to $0$ as violations emerge. In longer runs,
\texttt{causality\_health} may later recover if the violation rate
falls below the rolling evaluation window, reflecting the dynamic
nature of effective skew over time.
This demonstrates that real AI inference does not inherently cause
causal inconsistency.

\subsection{Skew Sweep and Threshold Behavior}
We perform an updated skew sweep focused on the low-millisecond onset
region and selected higher-skew conditions. In synchronized runs, no
violations are observed at 0\,ms, 1\,ms, 2\,ms, or 3\,ms. Clear
violations emerge at 5\,ms and above, placing the onset region between
3\,ms and 5\,ms in the current setup. At higher skew values, throughput
remains largely unchanged while negative spans and causality-health
degradation become pronounced. In longer runs, however, violation
behavior is not strictly monotonic: negative span rates can stabilize
and, in some cases, decrease over time, suggesting that effective skew
evolves dynamically due to relative clock drift between nodes.

\begin{figure}[!htbp]
  \centering
  \includegraphics[width=0.55\linewidth]{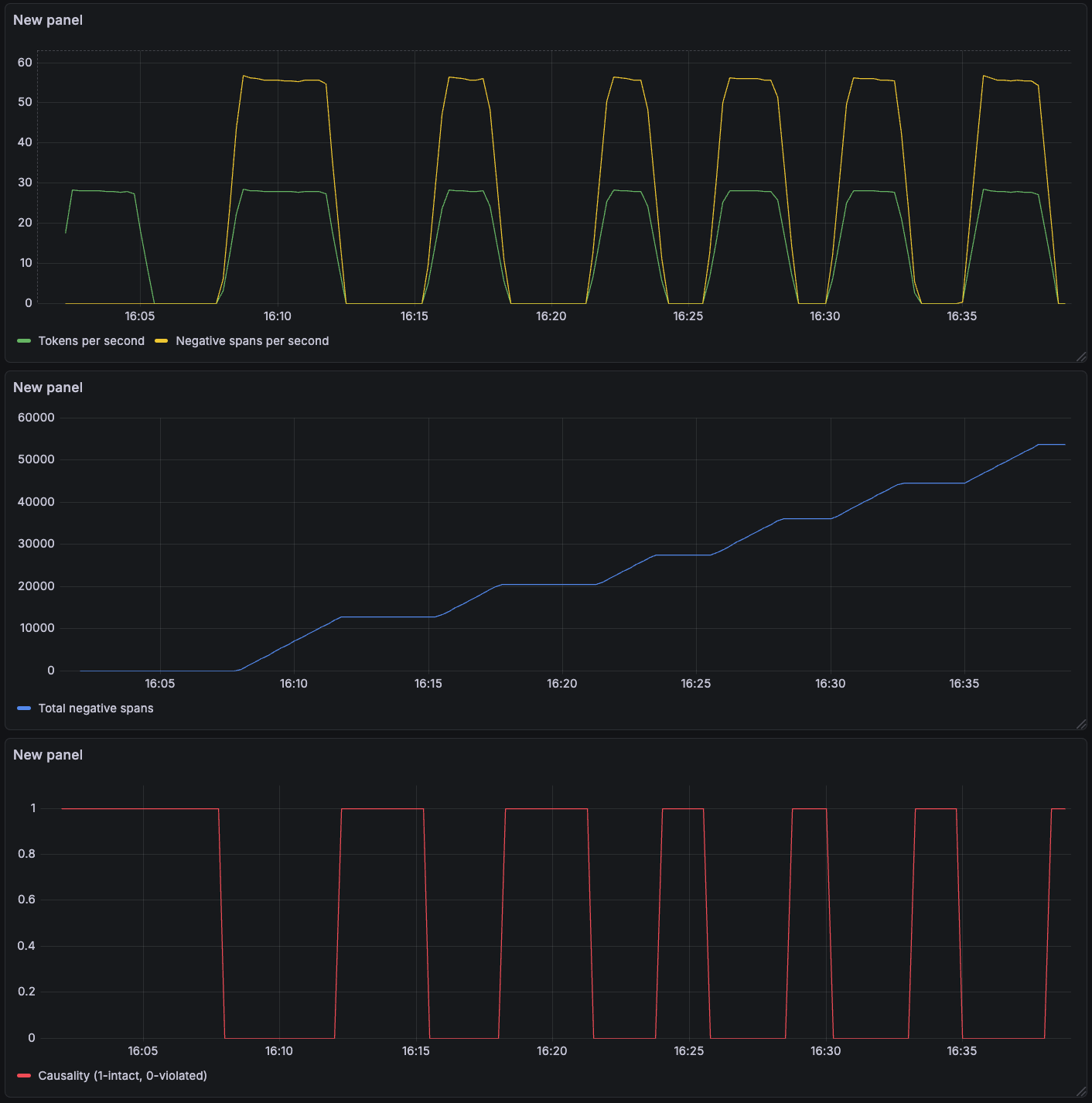}
  \caption{Skew sweep results. No violations are observed under
    synchronized conditions through 3\,ms skew, while clear violations
    appear by 5\,ms. This places the onset region between 3\,ms and
    5\,ms in the current setup. At higher skew values, violations
    become substantial, although longer runs show that violation
    behavior may evolve dynamically rather than increase monotonically
    forever.}
  \label{fig:skew-sweep}
\end{figure}

Figure~\ref{fig:skew-sweep} summarizes violation behavior across the
updated skew sweep. The transition from zero violations through
3\,ms to clear violations at 5\,ms indicates threshold-like behavior
within the tested skew resolution.

\subsection{Causality Health Signal}
The \texttt{causality\_health} metric remains healthy under zero skew
and in synchronized low-skew runs through 3\,ms. Once skew reaches the
onset region and beyond, the metric transitions to unhealthy as
violations emerge. In longer runs, \texttt{causality\_health} can
recover if the violation rate later falls outside the rolling
evaluation window, reflecting the time-varying nature of effective
skew.

\subsection{Dynamic Behavior and Summary of Experimental Observations}
Across the completed experiments, we observe the following behaviors:
\begin{itemize}[leftmargin=*]
  \item zero violations under zero skew,
  \item no violations observed through 3\,ms in synchronized runs, with
    clear violations emerging by 5\,ms,
  \item little to no impact on throughput or functional correctness, and
  \item violation behavior that can be strong but time-varying in
    longer runs, including cases where negative span growth slows or
    stops over time.
\end{itemize}
These observations hold across the completed Kafka and ZeroMQ
experiments. While repetition confirms the overall qualitative pattern,
some later exploratory runs were not normalized to identical durations,
so comparisons of absolute totals across all runs should be interpreted
cautiously. The strongest conclusions should therefore be taken from
the synchronized baseline, the low-skew runs through 3\,ms, and the
clear onset of violations by 5\,ms.

\section{Analysis}
\label{sec:analysis}

\subsection{What Breaks and What Does Not}
Notably, the following remain correct:
\begin{itemize}[leftmargin=*]
  \item Inference outputs.
  \item Message delivery and ordering.
  \item System throughput.
\end{itemize}
What breaks is the system's ability to infer and prove cause-and-effect
relationships from timestamps.

\subsection{Clock Skew as Trigger, Not Root Cause}
Clock skew triggers the failure, but the root cause is continued
reasoning about causality as if timestamps remain safe once skew
exists. Small skew is unavoidable; unsafe assumptions are not.

The later observation of self-recovery strengthens this interpretation.
Injected skew alone does not fully determine the observed failure
state; rather, the effective skew seen by the observability layer is
shaped by both the injected offset and the evolving relative drift
between nodes over time.

Importantly, improving clock synchronization accuracy alone only shifts
the effective clock error threshold $\varepsilon$ (the maximum
inter-stage clock skew); it does not eliminate the failure mode, which
arises from assuming timestamp-based causality remains safe once skew
exists.

\subsection{Queueing Effects as a Non-Causal Factor}
Queueing effects are considered as a potential confounding factor, as
increased backlog introduces additional positive delay between events.
However, queueing alone cannot produce negative timing spans or cause
timestamps to imply reversed causal orderings.

To validate that observed causality violations are not artifacts of
backlog, we analyze system behavior under controlled load conditions
where queueing delay is bounded. Under these conditions, violations
persist once clock skew is introduced, despite stable throughput and
limited queue growth.

Queueing theory provides intuition for backlog formation at the
inference stage. Let $\lambda$ denote the request arrival rate and
$\mu$ the effective service rate. When $\lambda < \mu$, the system
operates in a stable regime with bounded queueing delay. Our
experiments include runs where $\lambda$ is well below $\mu$, confirming
that causality violations occur even in the absence of significant
backlog.

These results demonstrate that queueing effects do not explain the
observed negative timing spans. Clock skew is the triggering factor,
while queueing only affects latency magnitude, not causal ordering.

\subsection{Error-Bound Interpretation of Timestamp-Based Causality}
\label{sec:error-bound}
Let $\varepsilon$ represent the maximum clock error between any two
stages. Timestamp-based causality remains valid only when $\varepsilon$
is smaller than the minimum inter-stage temporal separation used to
infer ordering. Once $\varepsilon$ exceeds this implicit error budget,
timestamps may imply contradictory orderings.

Formally, timestamp-based causal reasoning is safe only when the
following condition holds:
\begin{align}
  \varepsilon &< \Delta t_{\min}
    &&\Longrightarrow\quad \text{causality is preserved,}\\
  \varepsilon &\geq \Delta t_{\min}
    &&\Longrightarrow\quad \text{causality may be violated,}
\end{align}
where $\Delta t_{\min}$ represents the minimum observable time
difference between two causally related events such that the system's
observability logic can unambiguously infer their ordering using
timestamps (e.g., between a send and receive, or between upstream and
downstream stages).

In practice, $\Delta t_{\min}$ is not a single constant but exhibits
variability due to scheduling jitter, buffering, and instrumentation
resolution. Therefore, violation probability increases as
$\varepsilon$ exceeds increasing quantiles of the $\Delta t_{\min}$
distribution. Within the tested skew resolution, this behavior
manifests as threshold-like rather than gradual; however, finer-grained
skew sweeps would allow probabilistic characterization of the
transition region. In practice, violation frequency can be viewed as a
function of $\varepsilon$ relative to the empirical $\Delta t_{\min}$
distribution.

\paragraph{\texorpdfstring{(a) $\varepsilon < \Delta t_{\min}$ --- causality preserved.}{(a) epsilon < Delta t\_min --- causality preserved.}}
\begin{verbatim}
  Producer    Preproc    Inference    Postproc
   t=100       t=105       t=110        t=115
       -----------> -----------> ----------->
                Correct causal ordering
\end{verbatim}

\paragraph{\texorpdfstring{(b) $\varepsilon \geq \Delta t_{\min}$ --- causality violated.}{(b) epsilon >= Delta t\_min --- causality violated.}}
\begin{verbatim}
  Producer    Preproc    Inference    Postproc
   t=100       t=105       t=95         t=110
       -----------> -----------> ----------->
            X  Inference appears before cause
\end{verbatim}

In our experiments, $\Delta t_{\min}$ is empirically estimated as the
minimum observed positive inter-stage timestamp separation under
zero-skew conditions, between causally related events $E_i$ and $E_j$
across stages as recorded and propagated by the system.

In the current synchronized setup, no violations were observed through
3\,ms skew, while clear violations appeared by 5\,ms. This places the
empirical onset region between 3\,ms and 5\,ms for the present
pipeline. The later recovery behavior seen in longer runs also suggests
that $\varepsilon$ should be treated as a time-varying quantity shaped
by both injected skew and relative drift, rather than as a fixed static
value equal only to the application-level offset.

Finer-grained skew sweeps, especially around 4\,ms, may further refine
the onset boundary and distinguish whether the observed transition is
sharp or probabilistic in the current setup.

\clearpage
\section{Real-World Application Impact}
\label{sec:impact}

The failure mode demonstrated in this work is not theoretical. It has
direct implications across several real-world AI application domains
where causal correctness, explainability, and auditability are
essential.

\subsection{Multi-Tenant AI Inference Platforms}
Modern AI inference platforms frequently multiplex requests from many
tenants onto shared infrastructure~\cite{akoush2022,modserve,priority}.
Timestamp-based observability is used to enforce fairness, attribute
resource usage, debug tail latency, and support billing or chargeback
models. When causal timelines become inconsistent, these guarantees
silently erode. A tenant may appear to have received preferential
treatment, or a billing dispute may become impossible to resolve, even
though the underlying execution was correct. In such environments,
loss of causal trust undermines both operational confidence and
commercial accountability.

\subsection{Autonomous Agents and Control Loops}
AI systems increasingly operate as autonomous agents that invoke tools,
trigger downstream actions, and participate in closed-loop control
systems. Correct sequencing of observations and actions is critical for
safety, rollback, and explainability. If timestamps imply that an
action occurred before the signal that triggered it, post hoc analysis
becomes unreliable. While the agent's behavior may remain deterministic,
the system can no longer prove why a particular action was taken,
making safe automation and human oversight significantly harder.

\subsection{Financial and Regulated Decision Systems}
In financial services, healthcare, and other regulated domains,
decision systems must provide defensible audit trails. Regulators and
auditors often require reconstruction of event timelines to verify
compliance with policies and regulations. Timestamp-based causal
inconsistencies invalidate such reconstructions. Importantly,
regulators typically treat unverifiable correctness as incorrectness.
Thus, even when outcomes are correct, loss of causal observability can
translate directly into regulatory risk.

\subsection{Incident Response and Forensics}
During incident response, operators rely on logs and traces to
reconstruct what happened, in what order, and why~\cite{dynacausal}.
Silent causal failures increase mean time to resolution by forcing
engineers to reason through contradictory evidence. Worse, once the
timeline is corrupted, it cannot be repaired retroactively. This
increases operational cost, erodes trust in observability tooling, and
complicates blameless postmortems.

The experiments in this paper show that such forensic degradation can
begin even while the system remains functionally healthy, making
observability failure especially dangerous during incident response.

\subsection{Distributed GPU Training and Collective Communication Fault Analysis}
In large-scale GPU training, collective operations like AllReduce,
AllGather, and ReduceScatter require all participating GPUs to
synchronize and exchange data in lockstep. Libraries such as NCCL,
Gloo, and MPI coordinate these across thousands of GPUs, and at that
scale, failures are routine. The difficulty is in locating the root
cause. Collective communication creates dependency chains where one
stalled GPU causes its peers to block, which causes their peers to
block, until eventually a GPU several hops away reports a timeout. The
symptom shows up far from the source. Tracing it back requires knowing,
with confidence, when each collective began, which GPUs participated,
which were late or absent, and where in the operation the chain first
broke. This is fundamentally an observability problem, and it depends
on accurate global time ordering. With it, automated pipelines can
pinpoint the faulty GPU, pull it from the topology, bring in a
replacement, and resume training from checkpoint. Without it, those
same pipelines either misidentify the culprit or fail to converge at
all.

\section{Design Implications and Solution Directions}
\label{sec:design}

The goal of this work is not to prescribe a single architectural fix,
but to motivate a shift in how distributed AI systems reason about
time and causality. Based on our findings, we outline several
complementary design directions.

\subsection{Detection and Signaling of Causal Trust}
The most immediate implication is the need for explicit detection of
when timestamp-based causality assumptions are violated. The causality
health signal introduced in this work provides a simple but powerful
mechanism for systems to declare when their internal timelines are no
longer trustworthy. Such signals can be used to gate automation,
trigger alerts, or annotate observability data to prevent incorrect
interpretation. In practice, causality health signals can be
interpreted as runtime indicators that $\varepsilon$ has exceeded
$\Delta t_{\min}$, allowing systems to stop trusting timestamp-based
reasoning before incorrect conclusions are drawn.

When \texttt{causality\_health} is zero, systems should treat
timestamp-based observability as untrustworthy and either gate
automation, annotate outputs, or fall back to alternative ordering
mechanisms rather than continuing to reason over potentially invalid
timestamps. Such mitigation mechanisms shift systems from implicit
timestamp trust to bounded or logically enforced ordering guarantees.

\subsection{Relation to Existing Causality and Time Semantics}
Prior work has proposed alternatives to pure timestamp-based
causality, including Hybrid Logical Clocks (HLC), physical-with-causality
(PWC) clocks~\cite{pwc}, and verifiable logical clocks such as
Chrono~\cite{chrono}. These approaches explicitly combine physical
time with causal ordering guarantees or provide mechanisms to detect
inconsistencies when clock uncertainty exceeds safe bounds. Similarly,
uncertainty-bounded time services such as TrueTime~\cite{spanner}
expose time intervals rather than point estimates, and production
tracing systems~\cite{dapper,otel} employ skew correction and
compensation heuristics to mitigate timestamp inconsistencies. Recent
work on clock-dynamics-aware anomaly detection~\cite{jamshidi2026}
further highlights that timing integrity itself is an active research
area.

Our work is complementary to these efforts. Rather than proposing a
new causality mechanism or evaluating existing alternatives, we
empirically demonstrate the failure modes that arise when systems
assume timestamps are always safe to reason with. The results motivate
the need for hybrid or uncertainty-aware causality mechanisms in AI
inference pipelines, particularly when automation and auditability
depend on correct causal interpretation.

Recent work such as LatencyPrism~\cite{latencyprism} focuses on shaping
and controlling inference latency to meet service-level objectives in
large language model inference systems. This work addresses a
complementary but orthogonal concern: the correctness of causal
observability under clock skew, even when latency and throughput
remain within targets. Together, these results suggest that meeting
latency SLOs does not imply that timestamp-based observability remains
trustworthy.

\subsection{Adaptive System Behavior}
Rather than assuming a single global mode of operation, AI systems can
adapt their behavior when causal trust is lost. Examples include
switching to conservative execution modes, introducing human-in-the-loop
review, or delaying irreversible actions until causal confidence is
restored. Such adaptations acknowledge that time-based reasoning is a
bounded assumption, not an absolute truth.

\subsection{Measurement and Validation of Time Error}
Independent measurement of time error, drift, and step changes remains
valuable. However, this work shows that being within nominal
synchronization specifications does not guarantee causal safety.
Measurement should therefore be paired with system-level validation
that explicitly tests causal assumptions under realistic skew
scenarios.

Our later experiments reinforce this point operationally. Large
uncontrolled infrastructure offsets can easily exceed the observed
causality-safe region unless clocks are explicitly validated and
corrected before measurement. Time validation must therefore be
treated as part of observability validation, not as a separate
infrastructure hygiene task.

\subsection{Practical Guidance for Deployment}
The findings suggest several practical steps for operators of
distributed AI inference systems.
\begin{description}[leftmargin=*,style=unboxed]
  \item[Estimating $\Delta t_{\min}$.] $\Delta t_{\min}$ can be
    empirically estimated by measuring the minimum positive
    inter-stage timestamp separation under zero-skew baseline
    conditions across representative workloads. Because
    $\Delta t_{\min}$ exhibits variability, lower quantiles of its
    observed distribution should be considered when determining safe
    skew bounds.
  \item[Setting alert thresholds.] Clock skew alert thresholds should
    be configured conservatively below lower quantiles of the
    observed $\Delta t_{\min}$ distribution. Where measured skew
    approaches these bounds, systems should treat timestamp-based
    observability as potentially unsafe.
  \item[Retrofitting existing tracing stacks.] Existing tracing
    systems can incorporate causality health signals by augmenting
    timestamp comparison logic with skew uncertainty bounds, hybrid
    physical-logical clocks, or uncertainty-aware ordering overlays.
\end{description}

\section{Threats to Validity and Limitations}
\label{sec:threats}

Several factors may influence the generality of the results presented
in this paper. First, the experiments use a representative but
specific AI inference pipeline rather than an exhaustive survey of all
possible architectures. Second, inference operates in a
generate-then-emit mode rather than true token-level streaming. This
choice reduces temporal resolution at the inference boundary and may
mask finer-grained causal effects that would be observable with
per-token timestamping. True streaming inference is expected to reduce
the effective $\Delta t_{\min}$ and enable more precise
characterization of causality thresholds and is therefore an important
direction for future work. Third, clock skew is injected at the
application level rather than through direct manipulation of operating
system clocks. Fourth, experiments are conducted on commodity hardware
within a controlled environment. Finally, the messaging substrate
represents a common class of forward-in-time-only systems but is not
the only possible transport.

These choices were intentional to isolate causal observability effects
under realistic but controlled conditions. Future work will expand
along these dimensions.

This study focuses on step-change skew and generate-then-emit
inference. Future work includes drift experiments, streaming inference,
nanosecond-resolution timestamps, and broader transport validation.
The completed transport results in this paper are based on Kafka and
ZeroMQ; Aeron remains under active exploration and is not yet part of
the completed validation set.

Some later exploratory runs were executed with unequal durations and
are therefore interpreted qualitatively rather than as directly
normalized rate comparisons. The strongest quantitative conclusions in
this paper are based on the synchronized baseline runs, the clean
no-violation runs through 3\,ms, and the clear onset of violations by
5\,ms.

The observation of self-recovery in longer runs also indicates that
effective skew is dynamic. This means that fixed application-level
skew injection should not be interpreted as the sole determinant of
long-run observability behavior in real systems where clocks continue
to drift.

We plan to release the reference implementation and experiment scripts
to support reproducibility and community validation.

\section{Conclusion}
\label{sec:conclusion}

This paper demonstrates that small clock skew can silently break
timestamp-based causal observability in distributed AI inference
systems even when functional behavior remains correct. In the
synchronized setup evaluated here, no violations were observed through
3\,ms skew, while clear observability failures emerged by 5\,ms,
despite stable throughput and correct outputs.

The completed transport results presented in this paper are based on
Kafka and ZeroMQ, both of which exhibit the same qualitative failure
mode. This indicates that the observed problem is fundamentally about
time alignment rather than a specific messaging stack. Aeron remains
under active exploration and is left for future completion.

The experiments also show that causality failure need not behave as a
permanently static state. In longer runs, violation rates can evolve
over time, including partial or full recovery, suggesting that
effective skew is shaped not only by injected offset but also by
relative clock drift between nodes.

Taken together, these results show that observability correctness
depends critically on time synchronization, and that distributed AI
systems may operate in a state where functionality appears normal
while observability is already causally invalid.

\bibliographystyle{plain}
\bibliography{refs}

\end{document}